\documentclass[runningheads]{llncs}

\usepackage{eccvabbrv}
\usepackage{graphicx}
\usepackage{booktabs}
\usepackage[accsupp]{axessibility}  % Improves PDF readability for those with disabilities.

\usepackage{hyperref}
\usepackage{orcidlink}
\usepackage{booktabs}
\usepackage{multirow}
\usepackage{array}
\usepackage{pifont}
\usepackage{enumitem}
\usepackage{graphicx}
\usepackage{wrapfig}
\usepackage{xcolor}

\begin{document}

% ---------------------------------------------------------------
\title{Timage: A Generative Text-in-Image Paradigm for Fine-Tuning Vision-Language Models}

\author{%
Yifeng Wu$^{1,2,3}$, 
Huimin Huang$^{3}$, 
Ruiluo Wu$^{2}$, 
Chunyi Lin$^{2}$, \\
Guanhua Chen$^{4}$ 
Xian Wu$^{3}$, Wang Song$^{2}$, Ruize Han$^{1}$\thanks{Corresponding authors.}}
\institute{%
$^1$Fudan University,
$^2$Shenzhen University of Advanced Technology,
$^3$Tencent Jarvis Lab,
$^4$Southern University of Science and Technology
}

\maketitle

\begin{abstract}
Multimodal Large Language Models (MLLMs) often lose track of the right image regions during fine-grained spatial reasoning, because a textual query rarely carries any explicit geometric anchor into the pixel domain. Prevailing remedies either rewire the model's weights or pad the prompt with verbose instructions, yet neither reliably pins the language to the correct visual coordinates without eroding the backbone's general competence. We introduce \emph{Timage}, a paradigm that recasts multimodal understanding as an alignment problem solved \emph{at the input}: the query is drawn, as a typeset overlay, onto the image itself. The placement and appearance of this overlay are produced by a \emph{Constrained Schr\"odinger Bridge} (cSB), an entropic optimal-transport sampler that factorizes layout synthesis into two coupled stochastic stages. The first stage, \emph{Region Search}, transports noise toward query-aligned image zones while obeying a hard occlusion barrier that protects salient foreground content; the second stage, \emph{Appearance Shaping}, sizes the glyphs through an ``ink-budget'' regularizer so that the rendered text stays legible and visually balanced. The resulting overlay behaves as an explicit attention beacon that channels the model's focus along spatial semantics. On the VMCBench suite, Timage paired with a modest 7B backbone clearly overtakes far larger proprietary systems as well as parameter-tuned baselines. The study positions deliberate input reconstruction as a powerful, architecture-neutral lever for strengthening multimodal reasoning.

\end{abstract}

\section{Introduction}
\label{sec:intro}
Multimodal Large Language Models (MLLMs) have reshaped how machines couple visual perception with linguistic reasoning, and they now span a wide task surface: answering questions about cluttered scenes, parsing documents, reading scientific charts, and more. Beneath these successes, however, lies a persistent structural mismatch--the query and the picture never truly share a representational footing. A textual instruction (a question, a directive, a reasoning cue) is tokenized into a discrete symbol stream, whereas the image enters as a dense, continuous tensor. The two only meet later, through implicit feature blending inside cross-attention layers. As a result, the network is forced to silently infer a mapping from abstract phrases--``to the left of,'' or a positional reference inside a packed layout--onto concrete pixel regions. When the task demands fine spatial discrimination, dense text reading, or chained logical steps, this inference frequently slips, and the model's attention drifts away from the region the query actually concerns.

Existing attempts to close this gap tend to fall into two families, each with its own ceiling. The first family, \emph{weight-space adaptation}, is typified by parameter-efficient fine-tuning such as LoRA~\cite{hu2022lora} and adapter modules~\cite{chen2022vision,sung2022vl,zhang2021tip,gao2024clip,hu2023llm}. These methods bend the model's internal parameters toward a task distribution. They work well inside a fixed domain, but the price is steep: the pre-trained model's breadth degrades, foundational knowledge is partially overwritten, robustness to distribution shift weakens, and both compute and storage grow as task variants accumulate. In effect, a representational defect is patched by overfitting parameters rather than by repairing the signal that enters the model. The second family, \emph{input-space prompting}, led by Visual Prompt Tuning (VPT), optimizes continuous learnable vectors appended to the input. Yet these vectors are semantically opaque perturbations. Even geometry-aware variants only reshuffle where abstract feature tokens sit; they never inject readable content. Such prompts therefore cannot tap the backbone's latent character-recognition skill or its pre-aligned vision-language grounding, because the prompt itself means nothing in human terms.

\begin{figure}[t]
\centering
\includegraphics[width=\linewidth]{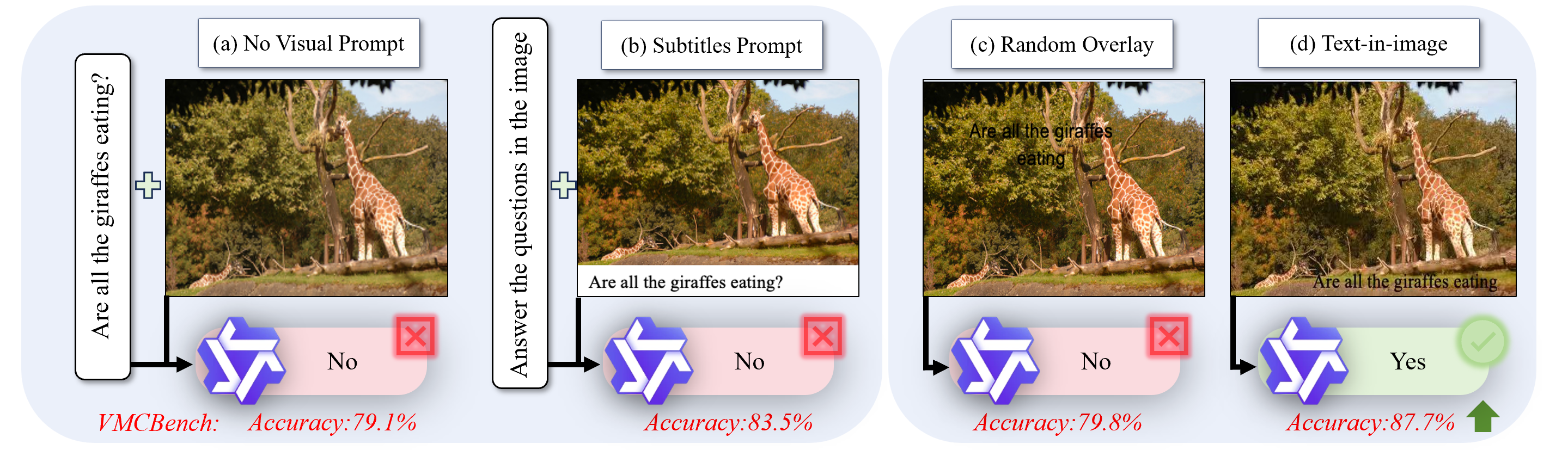} \vspace{-20pt}
\caption{
    (a-c) Prior approaches stumble on feature-level decoupling, spatial misplacement, or visual occlusion. (d) Timage instead synthesizes query-conditioned semantic overlays via the cSB sampler. By binding the instruction to the image manifold at the input stage, Timage lifts complex spatial reasoning and returns better-grounded multimodal answers.
}
\label{fig:example} \vspace{-15pt}
\end{figure}

These observations expose a missing piece in multimodal representation learning: there is no mechanism that explicitly turns the query text into a harmonious cross-modal cue co-located with the visual signal. We fill this gap with \emph{Text-in-Image} (Timage), which reframes multimodal reasoning as input-level alignment on a shared manifold. Rather than feeding the query as an external token stream, Timage paints the natural-language instruction onto the image as a typeset overlay, producing a query-conditioned visual prompt (Figure~\ref{fig:example}(d)). As a paradigm, Timage offers three distinguishing properties.
\ding{202} In contrast to the standard recipe of Figure~\ref{fig:example}(a), which embeds text and vision through separate branches, Timage \textbf{routes everything through a single visual input}.
\ding{203} Timage acts as an \textbf{external adapter that leaves the MLLM untouched}.
\ding{204} Timage is \textbf{simultaneously human-legible and machine-actionable}: the cross-modal cue is literally visible, which is also a step toward interpretable multimodal AI.

Realizing this is not trivial. As Figure~\ref{fig:example}(b-c) shows, naively stamping the text as a fixed caption~\cite{wang2025cure} or at a random spot is either too rigid or too unstable. We therefore cast the problem generatively, as a \emph{Constrained Schr\"odinger Bridge} (cSB). We split text-in-image inlaying into two questions--\emph{where} to write and \emph{how} to write--and solve them in sequence. For \emph{where}, we pose two objectives: pull the writing region toward query-relevant content (a low semantic potential) while honoring a hard prior that no salient foreground object may be covered. For \emph{how}, we auto-tune glyph size through an ``ink usage'' budget so the text fits gracefully. Both objectives become control conditions that steer a Schr\"odinger-Bridge diffusion process generating the overlay layer step by step.

We benchmark Timage on VMCBench~\cite{zhang2025automated}, which consolidates 20 diverse VQA datasets. With only a 7B backbone, Timage reaches a state-of-the-art mean accuracy of 87.7\%. It clears massive proprietary systems by wide margins (e.g., +7.4\% over GPT-4o and +2.7\% over Qwen2-VL-72B) and, strikingly, beats a full fine-tuning baseline by +2.5\%. These results argue that rebuilding the input can deliver more capability than adapting parameters. Cross-architecture tests confirm generality, with gains of +4.8\% to +8.6\% spanning backbones from LLaVA-1.5 to the recent Qwen3.
Our contributions are:
\begin{enumerate}
\item We present Timage, a paradigm that fuses linguistic semantics with the visual signal by rendering the query as an optimized overlay--single-modality at the input, plug-and-play without model edits, and visibly interpretable.
\item We cast overlay-layout synthesis as a Constrained Schr\"odinger Bridge, yielding a principled generative procedure whose energy functional trades off semantic pull against hard physical constraints.
\item We set new state-of-the-art numbers on VMCBench in a backbone-frozen, training-free regime, with stronger generalization and efficiency than both large proprietary models and parameter-efficient fine-tuning.
\end{enumerate}

\section{Related Work}

\label{sec:related_vlm}
\paragraph{Fine-Tuning of Vision-Language Models.}
Adapting Vision-Language Models (VLMs) is presently dominated by three parameter-efficient strands~\cite{han2024parameterefficientfinetuninglargemodels,ding2023parameter,xu2026parameter}. Adapter-style methods~\cite{sung2022vl,zhang2021tip,gao2024clip,hu2023llm,chen2022vision} graft small task-specific bottlenecks onto a frozen backbone to allow modular updates. Low-Rank Adaptation (LoRA) and its descendants~\cite{hu2022lora,liu2024dora,agiza2024mtlora,wang2024lora,liang2024inflora,zanella2024low,luo2023lcm} express weight updates through low-rank factors, trading expressive range for a small storage footprint. In parallel, Visual Prompt Tuning (VPT) and relatives~\cite{han20232vpt,das2023learning,huang2023diversity,sohn2023visual,gao2022visual} prepend learnable tokens to the input across transformer depths. Effective as they are at tuning weights or latent vectors, these techniques live almost entirely in latent feature space; their abstract codes resist human reading and can falter when the task needs tight alignment with explicit spatial-semantic constraints. Departing from such feature-centric edits, our work pursues a different route--steering model behavior by writing explicit semantic cues straight into the input pixels.

\label{sec:related_gen}
\paragraph{Generation in Service of Multimodal Understanding.}
Folding generative steps into the understanding pipeline has become a fertile direction. The literature ranges from textual Chain-of-Thought extensions~\cite{wei2022chain,feng2023towards,zhang2024chain} to visual-generative reasoning~\cite{tian2024visual,chen2020generative,van2016conditional,reed2017parallel,esser2021taming,van2017neural,razavi2019generating,yu2022scaling,yu2021vector}, where models emit intermediate sketches, heatmaps, or masks to scaffold hard logic. Other lines exploit interactive generation~\cite{jonathanho2020denoising,kodaira2025streamdiffusion,song2020denoising,song2020score,lu2022dpm,rombach2022high,luo2023lcm,shih2023parallel,du2020compositional}, producing iterative cues that sharpen spatial grounding and cross-modal alignment downstream. Most of this work, however, treats the reasoning trace as a side modality or a separate step kept apart from the original visual input. While that decoupling cleanly separates perception from reasoning, it can introduce a mismatch between the reasoning context and the primary scene. Motivated by this, we ask whether the reasoning prompt can be inscribed directly into the original image, yielding a self-contained, instruction-aware context.

\label{sec:related_sb}
\paragraph{Schr\"{o}dinger Bridges in Diffusion Models.}
The Schr\"{o}dinger Bridge (SB) framework has recently entered diffusion modeling as a way to solve entropic optimal-transport problems. This body of work includes theoretical advances~\cite{gu2022stochastic,fang2025neuralized,qiu2025finding} that characterize optimal stochastic paths between distributions, alongside practical deployments~\cite{su2022dual,kim2024latent,kim2023unpaired,liu20232,shi2023diffusion,tang2024simplified} in image translation and molecular conformer search. Newer studies push SB toward structured optimization~\cite{li2025audio,gushchin2024light,gushchin2024adversarial,kim2025discrete}, weaving geometric or physical priors into the diffusion to enable constrained synthesis. Extending SB to structured layout generation such as text overlays raises fresh difficulties: classical SB targets unconstrained transport, whereas a legible prompt must respect strict geometric and visibility limits so that essential image content is never occluded. Reconciling these competing objectives inside an SB process is still open, and we take a step toward it with a constrained generative formulation.

\section{Method}
\label{sec:method}
We propose Timage, a paradigm for visual-language interaction that writes the natural-language query into the image as a semantically aligned, spatially admissible overlay. Whereas conventional systems carry text as an external token sequence detached from image geometry, Timage produces a query-conditioned visual prompt through a controllable stochastic generator. This couples linguistic meaning with spatial context directly: the overlay nudges a downstream model toward task-relevant regions while a hard constraint forbids it from covering salient foreground content.

\begin{figure}[t]
\centering
\includegraphics[width=\linewidth]{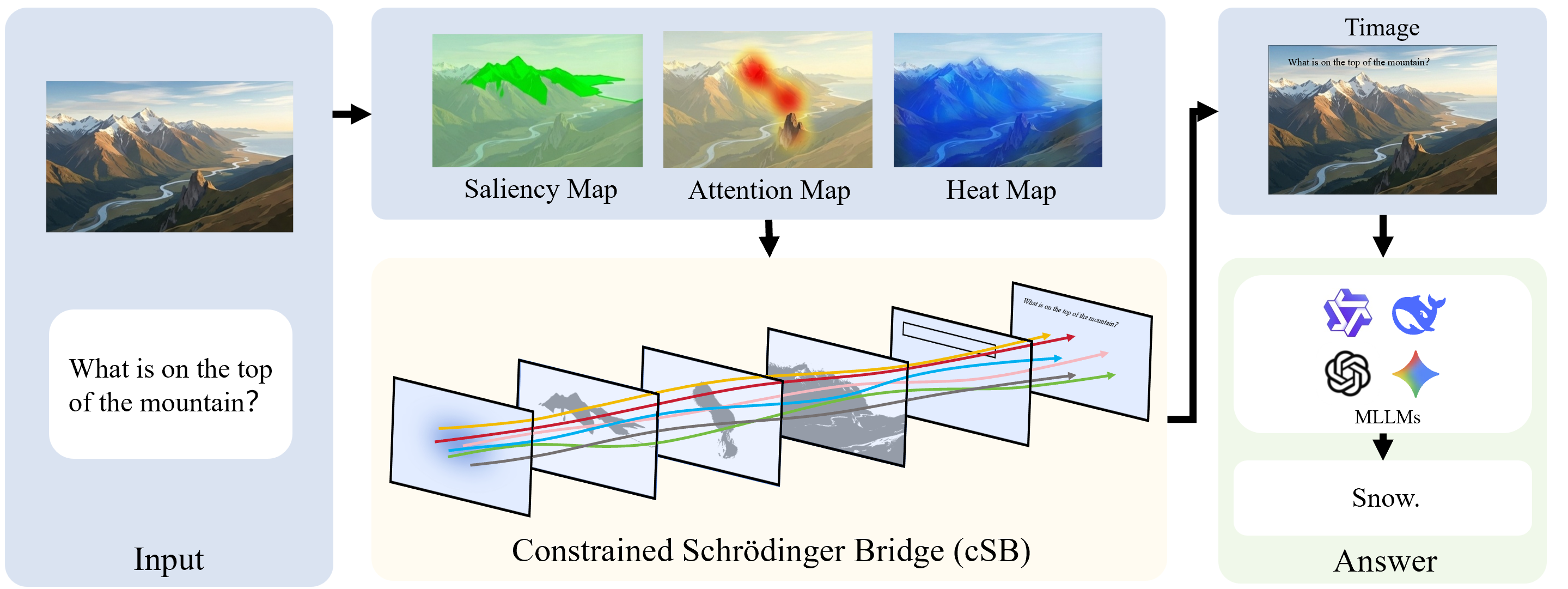} \vspace{-10pt}
\caption{
   Overview of Timage. From an image and its question we build a feasible manifold $\Omega$ through an Admissible Mask that merges semantic relevance with a hard non-occlusion prior. A Constrained Schr\"odinger Bridge then renders the inlaid text via a projected stochastic process, guaranteeing both geometric validity and downstream accuracy.
}
\label{fig:framework} \vspace{-15pt}
\end{figure}

\subsection{Overview and Notation}
\label{subsec:preliminaries}
Take an image $x_0$ and a query $q$; our aim is to inscribe $q$ onto $x_0$, much as one would letter a caption onto a painting. Two decisions govern the result: \emph{where} the text lives and \emph{how} it looks. We encode the first as a placement field and the second as a style field, and we score any candidate glyph layout $\mathcal{T}$ with a single \emph{writing energy}
\begin{equation}
    E(\mathcal{T}; x_0, q) = \lambda_{\text{zon}}\, E_{\text{zon}}\big(\alpha(\mathcal{T}); x_0, q\big) + \lambda_{\text{sty}}\, E_{\text{sty}}\big(\beta(\mathcal{T})\big),
    \label{eq:energy_func}
\end{equation}
where $\alpha \in [0,1]^{H \times W}$ is the placement field marking where $\mathcal{T}$ is written, and $\beta$ summarizes its appearance. The two terms read as:
\begin{itemize}[left=0pt]
    \item \emph{Placement energy} $E_{\text{zon}}$: rewards regions $\alpha$ that are query-relevant yet do not sit on top of the image content the query refers to.
    \item \emph{Style energy} $E_{\text{sty}}$: imposes appearance priors--an ink budget and spatial smoothness--so that $\mathcal{T}$ stays readable. Note that $\beta$ inherits constraints from the chosen placement $\alpha$.
\end{itemize}
Sections~\ref{subsec:problem_def} make $E_{\text{zon}}$ and $E_{\text{sty}}$ precise.

\paragraph{Why a distribution, not a point.}
As with lettering a painting, neither the region nor the style is uniquely determined: many placements and many sizes are acceptable. We therefore want a method that \emph{samples} from the set of admissible layouts rather than collapsing onto one. This is exactly what motivates the Schr\"odinger-Bridge formulation of Section~\ref{subsec:csb_framework}.

\subsection{Building the Energy Function}
\label{subsec:problem_def}
Given the pair $(x_0, q)$, we synthesize $x_1 = x_0 \oplus \mathcal{T}$, where $\mathcal{T}$ renders the content of $q$. Writing $\mathcal{T}$ requires a placement field $\alpha(\mathcal{T})$ and an appearance field $\beta(\mathcal{T})$ subject to the following constraints.

\subsubsection{Placement Constraint $E_{\text{zon}}$.}
We want $\alpha$ (shorthand for $\alpha(\mathcal{T})$) to be \emph{near} the image evidence implicated by $q$ but \emph{off} of it. Proximity helps the MLLM associate the inscribed $\mathcal{T}$ with the relevant region; non-coverage protects the very content the query depends on, especially query-related foreground objects. We make this concrete in two parts.

\ding{202} \emph{Semantic pull (soft).}
To express soft alignment, we first form a query-relevance heatmap $H_q$. With $e_q = \text{Encoder}_{\text{text}}(q)$, we read cross-attention between $e_q$ and the visual patches of a frozen VLM (e.g., Qwen2.5-VL) to produce $H_q$. The placement $\alpha$ should overlap with the spatial focus indicated by $e_q$, giving
\begin{equation}
\label{eq:sem}
    E_{\text{sem}}(\alpha) = -\int \alpha(u)\, H_q(u)\, du,
\end{equation}
with $u$ a spatial coordinate; stronger overlap means lower energy.

\ding{203} \emph{Non-occlusion (hard).}
We deliberately treat occlusion as a barrier rather than a soft penalty. We assemble an \emph{Admissible Mask} $M_{\text{cand}} \in \{0,1\}^{H \times W}$ that delimits the feasible region by fusing object and saliency evidence:
\begin{equation}
    M_{\text{cand}}(u) = \sigma\!\left( \mathbf{I} - \Big( \bigcup_i O_i(u) \Big) - \gamma\, S(u) \right),
    \label{eq:safety_mask}
\end{equation}
where $\mathbf{I}$ is the all-ones $W\times H$ matrix, $O_i(u)$ is the instance mask of object $i$ from a segmenter (e.g., Mask R-CNN), $S(u)$ is a dense saliency map, $\gamma$ weights the saliency term, and $\sigma$ binarizes via thresholding. Non-occlusion is enforced \emph{hard}: we require $\alpha \subset M_{\text{cand}}$, so feasibility is guaranteed rather than merely encouraged.

\subsubsection{Style Constraint $E_{\text{sty}}$.}
With $\alpha$ fixed, we render $\mathcal{T}$ strictly inside it and shape its appearance through $\beta$. In practice the dominant knob is font size, which we govern with a \emph{Query-Adaptive Ink Budget}: just as physical writing consumes ink, we cap the rendered ink so glyph size adapts to the text and the scene. The target ink volume is
\begin{equation}
    \mathrm{ink}(\mathcal{T}) = \kappa \cdot \text{Length}(q) \cdot \text{Scale}(\alpha),
    \label{eq:dynamic_ink}
\end{equation}
where $\text{Length}$ counts tokens, $\text{Scale}$ is the resolution factor of the placement region, and $\kappa$ is the nominal ink per character per unit area.

\subsection{Constrained Schr\"odinger Bridge}
\label{subsec:csb_framework}
We synthesize the overlay through a \emph{Constrained Schr\"odinger Bridge} (cSB): an optimal stochastic transport from a noise prior $\mu_0 = \delta$ to the inscribed text $\mathcal{T}$.

\subsubsection{Constrained Dynamics.}
We first transport the \emph{placement field} $\alpha$, keeping the trajectory inside the feasible manifold $M_{\text{cand}}$ of Section~\ref{subsec:problem_def}. Unlike unconstrained SB, our energy target rewards only semantic alignment (Eq.~\eqref{eq:sem}); spatial feasibility is delegated to the membership constraint $\alpha \subset M_{\text{cand}}$ from Eq.~\eqref{eq:safety_mask}.

Following~\cite{song2020score}, generation obeys a drift-diffusion SDE driven by a time-dependent velocity field $v_\theta$:
\begin{equation}
    d\mu_t = v_\theta(t, \mu_t, \mathbf{c})\, dt + \sqrt{2\mathcal{F}(t)}\, dW_t, \quad t \in [0, 1],
    \label{eq:sde_continuous}
\end{equation}
where $\mathcal{F}(t)$ is a fixed noise schedule and $\mathbf{c}$ bundles the conditions of Section~\ref{subsec:problem_def}. Concretely $\mathbf{c} = \{x_0, q, H_q, M_{\text{cand}}\}$ gathers the original input, the query-semantic cue, and the admissible region; it acts as the control signal throughout generation.

To merge these cues, at each step the network forms the composite input $[\mu_t; \Phi(\phi(t)); \Psi(\mathbf{c})]$, where $\phi(t)$ are Fourier time features broadcast to spatial size by $\Phi$, and $\Psi(\mathbf{c})$ injects the semantic features $(e_q, H_q)$ through cross-attention with $e_q = \text{Encoder}_{\text{text}}(q)$. To keep the path inside $M_{\text{cand}}$, we discretize with a \emph{Projected Euler--Maruyama} scheme~\cite{pierret2016non}; each step of size $\Delta t$ reads
\begin{equation}
 \label{eq:sde_discrete}
    \mu_{t+\Delta t} = \mathcal{P}_{\Omega}\!\Big( \mu_t + v_t\, \Delta t + \sqrt{2\mathcal{F}(t)}\, \Delta W_t \Big),
\end{equation}
where $v_t = v_\theta(t, \mu_t, \mathbf{c})$ is the predicted instantaneous drift and $\mathcal{P}_{\Omega}$ projects back onto the feasible set $\Omega = M_{\text{cand}}$.

Equation~\eqref{eq:sde_discrete} cleanly separates two responsibilities: the conditioned drift $v_t$ learns the \emph{semantic direction}, while the projection $\mathcal{P}_{\Omega}$ enforces \emph{spatial feasibility}. Running this to completion yields $\mu_1$, a layout that satisfies both the semantic objective and the non-occlusion barrier; we set the placement field $\alpha = \mu_1$.

Next we generate the \emph{appearance field} of $\mathcal{T}$. During this stage the ink budget enters the style energy $E_{\text{sty}}$ as a regularizer; tuning $\lambda_{\text{sty}}$ penalizes deviation so that the activated pixel count $\|\mathcal{T}\|_1$ drives toward $\mathrm{ink}(\mathcal{T})$. The glyph size thus adapts automatically--long queries draw a larger ink allotment (and may wrap to multiple lines), short queries stay compact--so the text settles naturally into the available space without crowding. Given region and size, line breaking follows automatically. Font and color are likewise carried through a Projected Euler--Maruyama solver: each iteration computes an intermediate noisy state seeded from default font and color values, which the drift then refines.

\subsubsection{Training Objective.}
We learn $v_\theta$ with a two-part loss, $\mathcal{L}_{\text{total}} = \mathcal{L}_{\text{cSB}} + \lambda\, \mathcal{L}_{\text{task}}$, jointly fitting the constraint energy and the downstream task.

\noindent\textit{Constrained flow-matching term ($\mathcal{L}_{\text{cSB}}$).}
Assuming a straight interpolation between $\mu_0$ and $\mu_1$, this term regresses the velocity onto the constant transport vector
\begin{equation}
    \label{eq:loss_cfm}
    \mathcal{L}_{\text{cSB}} = \mathbb{E}_t \Big[ \big\| v_\theta(t, \mu_t, \mathbf{c}) - (\alpha^* - \mu_0) \big\|_2^2 \Big],
\end{equation}
where $\alpha^*$ is a pseudo-target for the placement field. Since the placement has no unique optimum, we obtain $\alpha^*$ by \emph{approximate sampling} from the energy $E_{\text{zon}}$ to build an empirical distribution $\tilde{\mu}_1$; we use Langevin dynamics~\cite{gillespie2000chemical} for this, with details in the supplement.

\noindent\textit{Task-guided term ($\mathcal{L}_{\text{task}}$).}
To maximize downstream accuracy (e.g., VQA), we draw supervision from a frozen MLLM $\mathcal{M}$. With the rendered composite $x_1 = x_0 \oplus \mathcal{T}$ and query $q$, we minimize the negative log-likelihood
\begin{equation}
    \mathcal{L}_{\text{task}} = -\log P\big(y_{\text{gt}}, y_{\text{pred}} = \mathcal{M}(x_1)\big).
    \label{eq:loss_task}
\end{equation}
Gradients flow back through the discrete trajectory of Eq.~\eqref{eq:sde_discrete} using a straight-through estimator (STE)~\cite{bengio2013estimating} to bypass the non-differentiable projection, enabling end-to-end refinement of $v_\theta$ so that the layout is not only geometrically valid but also semantically optimal for the task.

\noindent\textit{Remark on the hybrid loss.}
The two terms are complementary. The task-guided term is direct: it back-propagates the downstream signal to improve placement and style. The constraint term is indispensable too: it supplies a placement prior that sharply shrinks the search space during diffusion and speeds convergence.

\subsection{Implementation Details}
We train for 100 epochs on 16 NVIDIA H20 GPUs (batch size 128) with AdamW ($lr=10^{-4}$, weight decay $10^{-2}$) and cosine annealing. The ink-budget scale is $\kappa=80$ pixels/character. To dampen layout bias and sampling variance, we apply self-consistency voting: we run $K$ independent inferences per image-query pair and select the final layout by majority vote, with $K=5$ in all experiments. Timage is built on PyTorch 2.1, Diffusers 0.24.0, and Qwen2.5-VL.

\section{Experiments}
\subsection{Datasets and Protocol}
\label{subsec:datasets}
We evaluate on VMCBench, which folds 20 diverse VQA datasets (e.g., VQAv2, MathVista, DocVQA) into one multiple-choice format of \textbf{9,450} items. The suite covers four axes--\emph{General Scene Understanding}, \emph{Complex Reasoning}, \emph{Text-Rich \& OCR}, and \emph{Structured Document Understanding}--probing robustness under varied spatial and semantic load. Using the unified format, we report Accuracy as the main metric for stability. For each item Timage samples $K$ overlays and votes for the final answer. We present Overall Accuracy and category-wise scores. All runs follow a strict zero-shot protocol with no task-specific fine-tuning, treating cSB as a plug-and-play, inference-time visual-prompting module on frozen MLLMs.

\subsection{Comparison with Baselines}
Table~\ref{tab:main_results} shows that \textbf{Timage} holds a clear, uniform edge across every visual domain, attesting to the robustness of cSB. Despite a compact 7B backbone, Timage reaches a state-of-the-art mean of \textbf{87.7\%}, edging past efficient peers such as Qwen3-8B (87.3\%) and decisively beating massive proprietary systems like GPT-4o (+7.4\%) and Qwen2-VL-72B (+2.7\%). The jump is sharpest where precise spatial alignment matters most: \textbf{Reasoning} (+7.6\%) and \textbf{Doc\&Chart} (+6.1\%), where query-conditioned overlays shrink the attentional search by physically tying intent to relevant regions while avoiding occlusion. Gains also carry to text-dense and general perception, with new bests in \textbf{OCR} (98.5\%, +3.2\%) and \textbf{General} (+5.7\%). Together these results indicate that inscribing the query into the visual geometry through our Constrained Schr\"odinger Bridge yields a stronger reasoning signal than simply enlarging the parameter count, bridging language and vision across domains.

\begin{table}[t]
\centering
\scriptsize
\caption{Performance comparison on VMCBench. {Timage} denotes our framework applied to \textit{Qwen2.5-VL-7B}, with ensemble voting ($K=5$).} \vspace{-10pt}
\label{tab:main_results}
\resizebox{\linewidth}{!}{%
\begin{tabular}{lccccc}
\toprule
\textbf{Model} & \textbf{General} & \textbf{Reasoning} & \textbf{OCR} & \textbf{Doc\&Chart} & \textbf{Avg.} \\
\midrule
\multicolumn{6}{l}{\textit{Proprietary / Large-Scale Models (>30B)}} \\
Qwen2-VL-72B      & 88.5 & 72.6 & 96.8 & 90.1 & 85.0 \\
GPT-4o            & 85.2 & 66.9 & 96.4 & 83.1 & 80.3 \\
Claude-3.5-Sonnet & 81.3 & 62.8 & 93.4 & 84.6 & 77.8 \\
Molmo-72B         & 82.9 & 66.6 & 94.7 & 81.1 & 78.7 \\
\midrule
\multicolumn{6}{l}{\textit{Open-Source Mid-Scale Models (10B--40B)}} \\
Qwen2.5-VL-14B   & 89.8 & 72.6 & 96.5 & 89.9 & 85.9 \\
Cambrian-34B      & 83.7 & 65.9 & 95.7 & 73.3 & 77.0 \\
VILA1.5-40B       & 82.5 & 65.3 & 93.2 & 67.4 & 74.7 \\
CogVLM2-19B       & 78.1 & 55.6 & 92.3 & 72.6 & 71.4 \\
\midrule
\multicolumn{6}{l}{\textit{Efficient Models ($\le$8B)} \& \textit{Our Method}} \\
Qwen3-8B          & 87.3 & 71.8 & 96.4 & 89.5 &  87.3 \\
Qwen2.5-VL-7B(Base) & 85.6 & 68.2 & 95.3 & 86.5 & 79.1 \\
Qwen2-VL-7B       & 84.5 & 62.7 & 96.4 & 80.1 & 78.1 \\
Molmo-7B-D        & 73.2 & 55.5 & 91.7 & 72.1 & 69.5 \\
Cambrian-8B       & 77.9 & 56.4 & 91.0 & 65.4 & 69.6 \\
Phi-3-Vision      & 74.1 & 56.4 & 90.6 & 73.8 & 70.3 \\
LLaVA1.5-7B       & 63.6 & 44.7 & 74.0 & 35.0 & 51.8 \\
\midrule
\textbf{Timage (Ours)} & \textbf{91.3} & \textbf{75.8} & \textbf{98.5} & \textbf{92.6} & \textbf{87.7} \\
\textit{\small (Base: Qwen2.5-VL-7B, K=5)} & \small{\textit{(+5.7)}} & \small{\textit{(+7.6)}} & \small{\textit{(+3.2)}} & \small{\textit{(+6.1)}} & \small{\textit{(+8.6)}} \\
\bottomrule
\end{tabular}%
}
\vspace{-0.2cm}
\end{table}

\subsection{Comparison with Alternative Strategies}
To probe Timage from every angle, we line it up against a broad slate of adaptation paradigms on the Qwen2.5-VL-7B backbone over VMCBench. Table~\ref{tab:comprehensive_comparison} groups the baselines into four families: (1) \emph{plain text prompting}, (2) \emph{heuristic spatial overlays}, (3) \emph{existing (noise-based) visual prompting}, and (4) \emph{parameter-efficient fine-tuning (PEFT)}. Notably, every baseline except plain text either needs task-specific gradient updates (PEFT, VPT) or leans on ad-hoc placement rules, whereas Timage runs strictly zero-shot and training-free, letting the Constrained Schr\"odinger Bridge produce semantic-aware layouts.

\begin{table}[t]
\centering
\caption{Comparison with representative fine-tuning paradigms on VMCBench.} \vspace{-0.2cm}
\label{tab:comprehensive_comparison}
\resizebox{\linewidth}{!}{%
\begin{tabular}{lccccccl}
\toprule
\textbf{Method} & \textbf{Category} & \textbf{Training} & \textbf{General} & \textbf{Reasoning} & \textbf{OCR} & \textbf{Doc\&Chart} & \textbf{Avg.} \\
 & & \textbf{Required?} & & & & & \\
\midrule
\multicolumn{8}{l}{\textit{Standard Baselines}} \\
Base Model (Text Only) & Text Prompt & No & 85.6 & 68.2 & 95.3 & 86.5 & 79.1 \\
+ Chain-of-Thought (CoT) & Text Prompt & No & 86.1 & 69.5 & 95.8 & 87.2 & 80.3 \\
\midrule
\multicolumn{8}{l}{\textit{Heuristic} \& \textit{Spatial Visual Prompts}} \\
Random Text Overlay & Heuristic Prompt & No & 86.9 & 69.5 & 95.8 & 86.1 & 79.8 \\
Top-Left Text Placement & Heuristic Prompt & No & 86.2 & 69.0 & 95.5 & 87.0 & 79.4 \\
Saliency-Based Placement & Heuristic Prompt & No & 86.1 & 68.5 & 96.0 & 88.2 & 80.5 \\
Bounding Box Prompt & Spatial Marker & No & 86.8 & 69.9 & 95.2 & 86.0 & 79.4 \\
\midrule
\multicolumn{8}{l}{\textit{Existing Visual Prompting}} \\
VPT-shallow\cite{jia2022visual} & Visual Prompt & Yes (Few-shot) & 86.8 & 70.8 & 96.1 & 88.4 & 81.9 \\
VPT-deep\cite{jia2022visual} & Visual Prompt & Yes (Few-shot) & 87.5 & 72.1 & 96.4 & 89.6 & 82.7 \\
CVPT\cite{huang2025cvpt}  & Visual Prompt & Yes (Few-shot) & 88.4 & 73.5 & 96.9 & 90.8 & 83.5 \\
\midrule
\multicolumn{8}{l}{\textit{Parameter-Efficient Fine-tuning}} \\
Adapter\cite{chen2022vision} & PEFT & Yes (Full) & 88.2 & 72.9 & 96.8 & 90.1 & 83.4 \\
LoRA\cite{hu2022lora} (Rank=64) & PEFT & Yes (Full) & 89.1 & 74.2 & 97.2 & 91.5 & 84.5 \\
Full Fine-tuning & Fine-tuning & Yes (Full) & 89.8 & 75.1 & 97.5 & 92.3 & 85.2 \\
\midrule
\textbf{Timage (Ours)} & \textbf{Text-in-Image} & \textbf{No} & \textbf{91.3} & \textbf{75.8} & \textbf{98.5} & \textbf{92.6} & \textbf{87.7} \\
\bottomrule
\end{tabular}%
}
\vspace{-0.2cm}
\end{table}

\paragraph{Semantics beat heuristics.}
Does merely stamping text on the image suffice? \emph{Random Text Overlay} (79.8\%) and \emph{Top-Left Placement} (79.4\%) barely move the needle over text-only (79.1\%), and \emph{Saliency-Based Placement} reaches only 80.5\%. \emph{Bounding Box Prompts}, carrying no semantic text, also stay below 80\%. So while location matters, \emph{semantic alignment} is decisive. Timage's \textbf{87.7\%} (+7.2\% over the best heuristic) confirms that cSB locates regions that are at once spatially admissible and semantically on-point--exactly where geometric rules fail.

\paragraph{Versus visual prompting.}
We inscribe \emph{natural language}, not \emph{noise vectors}. Constrained CVPT (83.5\%) improves on vanilla VPT (82.7\%) via geometry, yet still trails Timage by \textbf{+4.2\%}. The gap is telling: even optimal spatial constraints cannot supply the semantic grounding that hard reasoning needs. By writing the actual question into the visual manifold, Timage taps the MLLM's native OCR and grounding priors without any weight update--evidence that \textbf{semantic fidelity} matters more than abstract-representation tuning.

\paragraph{Versus parameter tuning.}
Strikingly, zero-shot Timage beats every PEFT variant and the \emph{full fine-tuning} baseline: +\textbf{3.2\%} over LoRA (84.5\%) and +\textbf{2.5\%} over Full FT (85.2\%) on average, with clear gains in Reasoning (+1.6\%) and Doc\&Chart (+0.3\%). Unlike PEFT and Full FT, which need task data and risk overfitting across heterogeneous distributions, Timage assembles an optimal visual context at inference. Reforming the input via cSB thus proves more effective than adapting parameters--efficient and broadly generalizable.

\subsection{Ablation Study}
\label{subsec:ablation}
Table~\ref{tab:ablation_study} isolates each component. From a random-placement base, adding Semantic Pull ($E_{\text{sem}}$) lifts accuracy by +2.5\%, confirming the value of query-relevant alignment. Activating the Admissible Mask ($M_{\text{cand}}$) gives the single largest gain (+3.0\%), underscoring that avoiding occlusion is critical in dense scenes. Style Regularization ($E_{\text{sty}}$) adds legibility and physical validity (+1.4\% combined). Finally, the Stochastic Ensemble ($K=5$) captures layout multi-modality for a closing +2.2\%, reaching 87.7\%.

\begin{table}[htbp] 
\centering
\scriptsize
\caption{Incremental impact of Timage components on VMCBench.} \vspace{-10pt}
\label{tab:ablation_study}
\begin{tabular}{lcccccc}
\toprule
\textbf{Configuration} & \textbf{Sem.} & \textbf{Mask}  & \textbf{Style} & \textbf{Ens.} & \textbf{Avg.} \\
 & ($E_{\text{sem}}$) & ($M_{\text{cand}}$) & ($E_{\text{sty}}$) & ($K=5$) & \\
\midrule
Random Placement (Base) & -- & -- & -- & -- & 79.1 \\
+ Semantic Pull         & \checkmark & -- & -- & -- & 81.6 \\
+ Admissible Mask       & \checkmark & \checkmark & -- & -- & 84.6 \\
+ Style Regularization  & \checkmark & \checkmark & \checkmark & -- & 86.0 \\
\textbf{Timage}    & \checkmark & \checkmark & \checkmark & \checkmark & \textbf{87.7} \\
\bottomrule
\end{tabular}%
\end{table}

\begin{figure}[t]
\centering
\includegraphics[width=\linewidth]{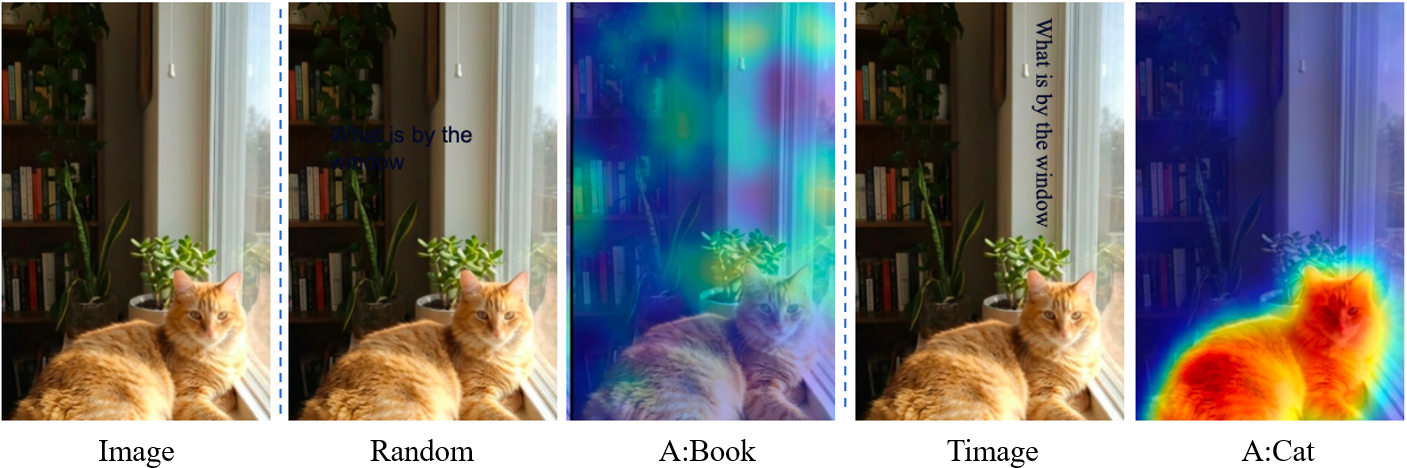} 
\caption{
    Qualitative comparison showing the benefit of Timage.
    Left: random question embedding covers irrelevant content and misleads the model to ``Book''.
    Right: cSB renders a semantically grounded, non-occluding overlay that preserves foreground saliency and guides attention to the window region, yielding the correct ``Cat''.
}
\label{fig:qualitative} 
\end{figure}

\subsection{Qualitative Results}
Figure~\ref{fig:qualitative} illustrates how Timage steers MLLM reasoning. The baseline's random inscription happens to cover key content and pulls attention toward an irrelevant background object (answering ``Book''). Timage instead lays down a semantically grounded, non-occluding overlay: by seating the query beside the target region it keeps foreground saliency intact and redirects the attention map to the correct object, so the model answers ``Cat''. This corroborates that an optimized text layout works as a potent visual prompt for accurate reasoning.

\section{Analysis}
\textbf{Generalization.}
\label{subsec:generalization}
To test universality, we attach Timage to five distinct backbones, from the classic LLaVA-1.5 to the recent Qwen3. Table~\ref{tab:generalization_models} shows uniform improvement, averaging \textbf{+6.1\%}. Even on the strong Qwen3-8B (87.3\% base), Timage climbs to \textbf{92.1\%} (\textbf{+4.8\%}), so the method does more than patch older models--it raises the ceiling of advanced VLMs. Gains concentrate on hard tasks, averaging \textbf{+6.8\%} in Reasoning and \textbf{+5.6\%} in Doc\&Chart. That this holds across CLIP-based, hybrid-ViT, and dynamic-resolution designs points to a shared bottleneck in current VLMs: the disconnect between linguistic queries and visual geometry. By rebuilding the input to align intent with spatial context, Timage behaves as a robust, model-agnostic, plug-and-play module that unlocks latent ability in any frozen MLLM.

\begin{table}[htbp] \vspace{-10pt}
\centering
\caption{Cross-architecture generalization of Timage.}  \vspace{-10pt}
\label{tab:generalization_models}
\resizebox{\linewidth}{!}{%
\begin{tabular}{lccccl}
\toprule
\textbf{Backbone Model} & \textbf{Base Acc.} & \textbf{+ Timage (Ours)} & \textbf{Improvement ($\Delta$)} & \textbf{Reasoning $\Delta$} & \textbf{Doc\&Chart $\Delta$} \\
\midrule
\textit{LLaVA-1.5-7B} & 51.8 & \textbf{57.4} & \textcolor{red}{\textbf{+5.6}} & +6.8 & +5.2 \\
\textit{Qwen2-VL-7B} & 78.1 & \textbf{83.9} & \textcolor{red}{\textbf{+5.8}} & +7.1 & +6.4 \\
\textit{InternVL-2-8B} & 72.5 & \textbf{78.3} & \textcolor{red}{\textbf{+5.8}} & +6.5 & +5.9 \\
\textit{Qwen2.5-VL-7B} & 79.1 & \textbf{87.7} & \textcolor{red}{\textbf{+8.6}} & +7.6 & +6.1 \\
\textit{Qwen3-8B} & 87.3 & \textbf{92.1} & \textcolor{red}{\textbf{+4.8}} & +5.9 & +4.5 \\
\midrule
\textbf{Average Improvement} & -- & -- & \textcolor{red}{\textbf{+6.1\%}} & \textbf{+6.8\%} & \textbf{+5.6\%} \\
\bottomrule
\end{tabular}%
}
\vspace{-0.2cm}
\end{table}

\begin{figure}[h] \vspace{-15pt}
\centering
\includegraphics[width=0.85\linewidth]{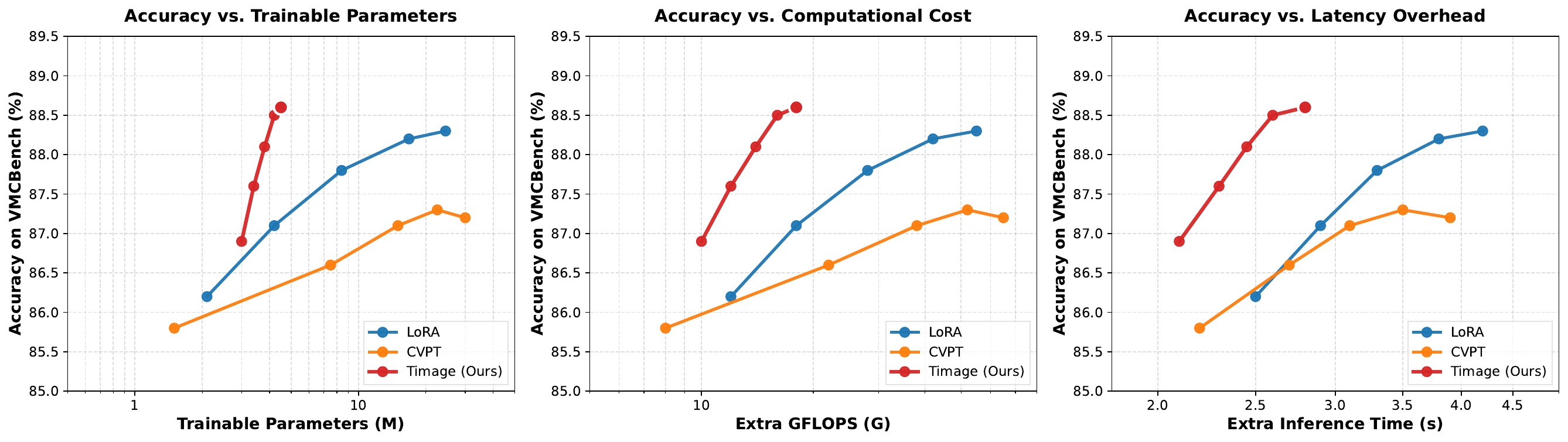} \vspace{-10pt}
\caption{
Efficiency-accuracy trade-offs on VMCBench.
}
\label{fig:eff} \vspace{-10pt}
\end{figure}

\textbf{Efficiency.}
To gauge parameter efficiency, we compare Timage with LoRA and CVPT across trainable-parameter budgets. As Figure~\ref{fig:eff} shows, Timage attains higher VMCBench accuracy with far fewer trainable parameters. Even scaled to 40M parameters, LoRA only inches to 88.3, while Timage keeps its lead at minimal cost--exposing the diminishing returns of brute-force parameter growth and reinforcing the value of our constrained, geometry-aware adaptation.

\begin{wrapfigure}{r}{0.5\textwidth} \vspace{-5pt}
    \centering
    \vspace{-20pt}
    \includegraphics[width=\linewidth]{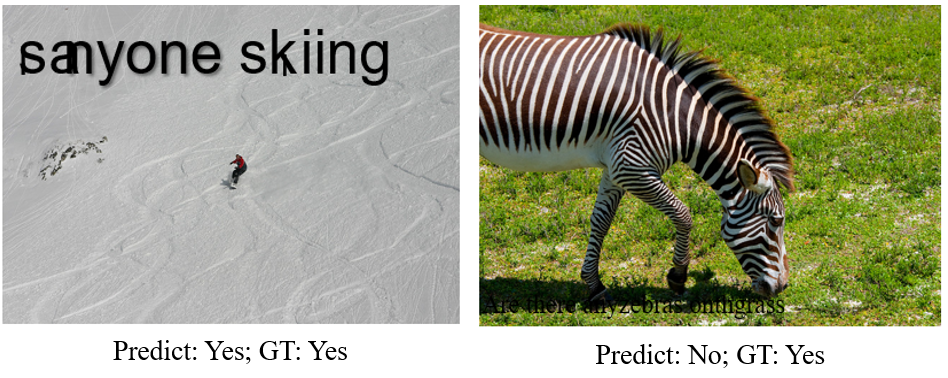} \vspace{-20pt}
    \caption{Failure cases of our method.}
    \label{fig:limations}
    \vspace{-25pt}
\end{wrapfigure}
\textbf{Limitations and Future Work.}
As Figure~\ref{fig:limations} indicates, while Timage strikes a good performance-efficiency balance, two minor artifacts surface in extreme cases. First, the rendered cue may occasionally be partially incomplete (e.g., a missing stroke); our experiments show this rarely dents reasoning accuracy. Second, in heavily textured or low-contrast areas the overlay's transparency can be suboptimal, letting the cue blend into the background. Future work will refine the renderer to ensure both the completeness and the distinctiveness of the cue without sacrificing efficiency.

\section{Conclusion}
We presented Timage, a paradigm that resolves the representational mismatch between linguistic queries and visual geometry by writing instructions onto the image as semantically grounded, spatially constrained overlays. Casting layout synthesis as a Constrained Schr\"odinger Bridge, Timage aligns intent with visual context while strictly preserving foreground saliency, removing the attentional drift built into token-decoupled designs. On VMCBench, this input-level manifold alignment surpasses both massive proprietary models and full fine-tuning, and it transfers robustly across heterogeneous backbones. The takeaway is that rebuilding the input signal through optimal transport can be a more efficient and potent path to multimodal reasoning than parameter adaptation--opening fresh directions for geometry-aware visual-language interaction.

% ---- Bibliography ----
\bibliographystyle{splncs04}
\bibliography{refs}
\end{document}